# Fuzzy finite element model updating using metaheuristic optimization algorithms

I. Boulkaibet[a] , T. Marwala[a], M.I. Friswell[b], H. Haddad Khodaparast[b] and S. Adhikari[b]

[a] Electrical and Electronic Engineering Department, University of Johannesburg, PO Box 524, Auckland Park 2006, South Africa
[b] College of Engineering, Swansea University Bay Campus, Crymlyn Burrows, Swansea SA1 8EN, United Kingdom

**ABSTRACT**

In this paper, a non-probabilistic method based on fuzzy logic is used to update finite element models (FEMs). Model updating techniques use the measured data to improve the accuracy of numerical models of structures. However, the measured data are contaminated with experimental noise and the models are inaccurate due to randomness in the parameters. This kind of aleatory uncertainty is irreducible, and may decrease the accuracy of the finite element model updating process. However, uncertainty quantification methods can be used to identify the uncertainty in the updating parameters. In this paper, the uncertainties associated with the modal parameters are defined as fuzzy membership functions, while the model updating procedure is defined as an optimization problem at each α-cut level. To determine the membership functions of the updated parameters, an objective function is defined and minimized using two metaheuristic optimization algorithms: ant colony optimization (ACO) and particle swarm optimization (PSO). A structural example is used to investigate the accuracy of the fuzzy model updating strategy using the PSO and ACO algorithms. Furthermore, the results obtained by the fuzzy finite element model updating are compared with the Bayesian model updating results.

.

## 1. Introduction

In many engineering areas, the finite element model (FEM) [1, 2, 3] is considered as one of the most practical numerical tools that can be used to obtain approximate solutions for real systems. Unfortunately, the accuracy of the models obtained by the FEMs degrades with the complexity of the modeled system. Furthermore, certain model parameters are uncertain and the variability of these parameters may reduce the accuracy of the FEM. Usually, the obtained finite element model is improved by correcting certain uncertain model parameters to reduce the error between the numerical model and the measured data. This process is known as “model updating” [4, 5]. Generally, there are two main approaches to perform a finite element model updating (FEMU): direct methods in which the FEM output is directly equated to the measured data [4]. In the second class, which is the class of the iterative approaches, an optimization procedure is performed to minimize the differences between the measured output and FEM output. In this optimization procedure, several unknown parameters are selected and corrected according to a preselected objective function [4].

In reality, the presence of uncertainties is inevitable in the modeling process as well as the experimental studies. These uncertainties can result from the mathematical simplifications that have been assumed during the modeling procedure (e.g., the variability of material-properties or mechanical joints). Furthermore, the experimental noises may contaminate the measured data where this kind of noise is irreducible (e.g., the accuracy of the equipment, such as the sensors, used to conduct an experimental study). These uncertainties may have a significant influence on the modeling process as well as the updating procedure. To obtain reliable models, the uncertainty quantification methods can be implemented to update FEMs [6].

Several uncertainty quantification methods have been used in FEMU such as: Bayesian method, perturbation methods and methods based on fuzzy sets [6, 7]. Generally, the uncertainty quantification methods are divided into two main classes:

probabilistic and possibilistic (non-probabilistic) methods [6]. In the probabilistic methods, the uncertain parameters are defined as random variables where the variability of each parameter is described by a probability density function (PDF). Both the Bayesian and the perturbation methods belong to the probabilistic class [6, 7, 8, 9, 10]. In the case that the density functions of the uncertain parameters are unknown (or their forms are unspecified), the possibilistic methods can be used to quantify the uncertainties associated with the structures. In this case, the variability of the uncertain parameters is described in a non-probabilistic way, e.g., intervals, membership functions… etc. The most recognized possibilistic methods are: the interval and the fuzzy logic methods. In the interval method, the uncertain parameters vary within intervals where each parameter is constrained by lower and upper bounds [6, 11, 12]. The interval finite element model updating concentrates on finding the minimal and maximal bounds of the unknown parameters. Khodaparast et al. [13] used the interval method to define the uncertainties associated with the measurements while a Kriging Meta model was employed to update the bounds of the unknown parameters. Describing uncertainties as fuzzy sets was first introduced by Zadeh [14]. In fuzzy finite element model updating (FFEMU) [6, 15, 16, 17], the fuzzy membership functions are used to define the uncertainties associated with the measured outputs (instead of probability density function) while the membership functions of the uncertain parameters are obtained by applying the interval finite element model updating at different membership function levels. Moens and Vandepitte [15] applied FFEMU to compute the uncertain frequency response functions of damped structures. Fuzzy sets were also used by Erdogan and Bakir [18] to represent the irreducible uncertainties caused by the measurement noise in the modal parameters while the membership functions of the uncertain parameters were obtained by minimizing an objective function using a genetic algorithm.

In this paper, a fuzzy finite element model updating approach based on metaheuristic optimization algorithms is discussed and used to quantify the uncertainty associated with the modal parameters. First, the FFEMU procedure is discussed in section 2 where the irreducible uncertainties are defined as fuzzy membership functions while the α-cut methodology is used to simplify the computing process by transferring the fuzzy calculations into a series of interval calculations. The objective function, which is defined by the upper and lower bounds of the numerical model and the measured data, is also presented in section 2 along with the interval natural frequencies and mode shapes of the numerical model. In section 3, the ant colony optimization (ACO) algorithm is discussed where this algorithm is used to compute the upper and lower bounds of the intervals at each α-cut. In section 4, the FFEMU based on ACO algorithm was investigated by updating a structure. The accuracy of the proposed FFEMU-ACO algorithm is highlighted by comparing its results with the results obtained by the FFEMU based on a PSO and Bayesian results. Section 5 concludes the paper.

## 2. Fuzzy finite element model updating

Instead of using the probability density functions to represent the irreducible uncertainty, the fuzzy finite element model updating method defines the uncertainty as a membership function. In general, the objective of the FEEMU method is to estimate the membership functions of the updated quantities given the membership functions of the measured outputs [6, 19]. This can be done by transferring the fuzzy arithmetic to a series of interval calculations. In this section, a brief introduction on fuzzy sets is given where the main idea of the α-cut technique will be presented. Next, the objective function used in this paper is defined in subsection 2.2, and since the interval technique is used to compute the membership functions, the bounds of the numerical eigenvalues are computed in the last subsection.

### 2.1 Introduction to fuzzy sets

Since the first introduction of the concepts of fuzzy logic in 1965 [14], the fuzzy methodology has been very attractive in many engineering areas such as: control systems and system identification. Unlike the standard set theory that separates between the members of the set and the non-members, a fuzzy set $\tilde{A}$ uses the degree of membership, which is measured by the membership function $\mu_{\tilde{A}}(x)$, to express the degree to which a member $x$ belongs to the fuzzy set $\tilde{A}$. A fuzzy set can be described as follows [18, 20]:

$$\tilde{A} = \{(x, \mu_{\tilde{A}}(x)) |\, x \in X, \mu_{\tilde{A}}(x) \in [0, 1]\} \tag{1}$$

As described in Eq. (1), $\mu_{\tilde{A}}(x) = 1$ indicates that $x$ is a member of the fuzzy set $\tilde{A}$, while $\mu_{\tilde{A}}(x) = 0$ indicates that $x$ does not belong to the set $\tilde{A}$. When $0 < \mu_{\tilde{A}}(x) < 1$, the element x belongs to the fuzzy set $\tilde{A}$ with certain degree $\mu_{\tilde{A}}(x)$. There are an infinite number of possible shapes in which a membership function can be defined, however the most frequent shapes are: trapezoidal, triangular and Gaussian shape. In this paper, the triangular form will be used to describe a membership function $\mu_{\tilde{A}}(x)$ such that:

$$\mu_{\tilde{A}}(x) = \begin{cases} 0, & x \le a, \\ \frac{x-a}{b-a}, & a < x \le b, \\ \frac{c-x}{c-b}, & b < x \le c, \\ 0, & c < x. \end{cases} \tag{2}$$

Usually, problems with fuzzy variables can be analyzed using the α-cut technique. In this technique, a membership function $\mu_{\tilde{A}}(x)$ of the member $x$ is divided into a series of sublevels. At each α-sublevel, the interval $\tilde{A}_\alpha = [\underline{x}, \overline{x}]$ is obtained where $\underline{x}$ and $\overline{x}$ are the upper and the lower bounds of the interval $\tilde{A}_\alpha$. The α-cut of the fuzzy set $\tilde{A}$ can also be described as follows:

$$\tilde{A}_\alpha = \{x \in X, \mu_{\tilde{A}}(x) \geq \alpha\} \tag{3}$$

In general, the problem of FFEMU can also be regarded as a propagation of a m-dimensional vector of fuzzy inputs $\tilde{\boldsymbol{x}}$, where the main idea is to attain the membership functions of the a d-dimensional vector of fuzzy outputs $\tilde{\boldsymbol{y}}$ when the membership functions of the input quantities $\tilde{\boldsymbol{x}}$ were given. The mathematical equation that describes this problem is given by: $\tilde{\boldsymbol{y}} = \boldsymbol{f}(\tilde{\boldsymbol{x}})$. Unfortunately, due to the complexity of this fuzzy problem, it cannot be numerically implemented [21]. The easiest way is to transfer the problem into an equivalent problem that can be solved as an interval analysis at different α-cuts [22]. Thus, the fuzzy uncertainty propagation is then expressed as the application of a series of interval analysis at a number of α-sublevels. There are two main strategies to evaluate fuzzy uncertainty propagation: interval algebra based approaches and global optimization based approach [21]. In the first approach, fuzzy variable is treated as an interval variable for each α-cut while the standard interval arithmetic will be used for the propagation procedure. In the second approach, which is the approach adopted in this paper, two optimization problems are solved for each α-sublevel. In this approach, the first optimization procedure is defined so that the minimum value of the output vector is obtained while the maximum value of the output vector is obtained from the second optimization procedure. Combining the results obtained by these two optimization problems for all α-sublevels, the membership functions of the output quantities are attained. The optimization problem at each α-level is described as follows:

$$\begin{cases} \overline{\boldsymbol{y}}_\alpha = \max\limits_{x_\alpha \in x_\alpha^I} \boldsymbol{f}(\boldsymbol{x}_\alpha) \\ \underline{\boldsymbol{y}}_\alpha = \min\limits_{x_\alpha \in x_\alpha^I} \boldsymbol{f}(\boldsymbol{x}_\alpha) \end{cases}, \alpha = \alpha^k, k = 1, \ldots, m \tag{4}$$

where $\overline{\boldsymbol{y}}_\alpha$ and $\underline{\boldsymbol{y}}_\alpha$ are the upper and lower bounds of the output vector, respectively. $\boldsymbol{x}_\alpha^I$ is the interval variable (defined earlier as $\tilde{A}_\alpha$). The optimization problem defined in Eq. (4) can be simplified to optimize a simple objective function. In the next subsection, the objective function will be described in details.

## 2.2 The objective function

In this paper, the FFEMU approach is defined as a global optimization based approach where an objective function is defined by a difference between the experimental and the numerical modal parameters. The objective function can be reformed to include both lower and upper bounds of the numerical and experimental modal parameters as described in Eq. (5) [18, 20]:

$$f(\boldsymbol{\theta}^I) = f(\overline{\boldsymbol{\theta}}, \underline{\boldsymbol{\theta}}) = \underline{\boldsymbol{e}}^T \boldsymbol{W}_l \underline{\boldsymbol{e}} + \overline{\boldsymbol{e}}^T \boldsymbol{W}_u \overline{\boldsymbol{e}} \tag{5}$$

where $\overline{\boldsymbol{\theta}}$ and $\underline{\boldsymbol{\theta}}$ are the lower and upper bounds of the interval updating parameters $\boldsymbol{\theta}^I$, respectively. $\boldsymbol{W}_l$ and $\boldsymbol{W}_u$ are the weighting matrices for the upper and lower parts of the objective function, respectively. $\underline{\boldsymbol{e}}$ represents the lower error where $\underline{\boldsymbol{e}} = \begin{bmatrix} \underline{\boldsymbol{e}}_f \\ \underline{\boldsymbol{e}}_\phi \end{bmatrix}$, $\underline{\boldsymbol{e}}_f = [\frac{\underline{\lambda}_1^m - \underline{\lambda}_1}{\underline{\lambda}_1^m}, \ldots, \frac{\underline{\lambda}_n^m - \underline{\lambda}_n}{\underline{\lambda}_n^m}]^T$ and $\underline{\boldsymbol{e}}_\phi = [\frac{\|\underline{\phi}_1^m - \underline{\beta}_1 \underline{\phi}_1\|}{\|\underline{\phi}_1^m\|}, \ldots, \frac{\|\underline{\phi}_n^m - \underline{\beta}_n \underline{\phi}_n\|}{\|\underline{\phi}_n^m\|}]^T$. $\underline{\lambda}_j^m$ and $\underline{\lambda}_j$ are the j-th lower bounds of the measured and the analytical eigenvalues, respectively. $\underline{\phi}_j^m$ and $\underline{\phi}_j$ are the j-th lower bounds of the measured and the analytical eigenvectors, respectively. $\underline{\beta}_j$ is a normalizing constant given by $\underline{\beta}_j = \frac{(\underline{\phi}_j^m)^T \underline{\phi}_j}{\|\underline{\phi}_j\|^2}$. In the same way, the error of the upper part of the objective function is given by $\overline{\boldsymbol{e}} = \begin{bmatrix} \overline{\boldsymbol{e}}_f \\ \overline{\boldsymbol{e}}_\phi \end{bmatrix}$, $\overline{\boldsymbol{e}}_f = [\frac{\overline{\lambda}_1^m - \overline{\lambda}_1}{\overline{\lambda}_1^m}, \ldots, \frac{\overline{\lambda}_n^m - \overline{\lambda}_n}{\overline{\lambda}_n^m}]^T$ and $\overline{\boldsymbol{e}}_\phi = [\frac{\|\overline{\phi}_1^m - \overline{\beta}_1 \overline{\phi}_1\|}{\|\overline{\phi}_1^m\|}, \ldots, \frac{\|\overline{\phi}_n^m - \overline{\beta}_n \overline{\phi}_n\|}{\|\overline{\phi}_n^m\|}]^T$. The upper bounds of j-th measured and the analytical eigenvalues are given by $\overline{\lambda}_j^m$ and $\overline{\lambda}_j$, respectively. $\overline{\phi}_j$and $\overline{\phi}_j^m$ are the j-th upper bounds of the analytical and measured eigenvectors, respectively. Again, the normalizing constant $\overline{\beta}_j$ of the j-th upper eigenvector is defined by $\overline{\beta}_j = \frac{(\overline{\phi}_j^m)^T \overline{\phi}_j}{\|\overline{\phi}_j\|^2}$. The optimization procedure defined by Eq. (5) is completed by adding two constraints to the problem. First, the updating parameter vector is bounded by the maximum and minimum values ($\boldsymbol{\theta}_{max}$ and $\boldsymbol{\theta}_{min}$) to keep the searching process within a limited region (which eventually limits the number of iterations). These maximum and minimum values help keeping the updating parameters physically realistic. Furthermore, to obtain convex membership functions, another constraints are included where the lower bounds $\underline{\boldsymbol{\theta}}^{k+1}$ of the updating parameters at $(k+1)$-th α-sublevel should be smaller than the previous $\underline{\boldsymbol{\theta}}^{k}$ that corresponds $(k)$-th α-sublevel. On the other hand, upper bounds $\overline{\boldsymbol{\theta}}^{k+1}$ of the updating parameters at $(k+1)$-th α-sublevel must be larger than the previous $\overline{\boldsymbol{\theta}}^{k}$ that corresponds $(k)$-th α-sublevel. Then, the new constraints are given in Eq. (6):

$$\begin{cases} \overline{\boldsymbol{\theta}}^{k} \leq \overline{\boldsymbol{\theta}}^{k+1} \leq \boldsymbol{\theta}_{max} \\ \boldsymbol{\theta}_{min} \leq \underline{\boldsymbol{\theta}}^{k+1} \leq \underline{\boldsymbol{\theta}}^{k} \end{cases}, \alpha = \alpha^{k}, k = 1, \dots, m \tag{6}$$

It is clear that at the first α-sublevel ( $\alpha^1 = 1$), which corresponds to $\mu_{\tilde{A}}(\theta_j) = 1, j = 1, \dots, d$, a simple deterministic optimisation is performed to obtain the center of membership functions. In the next subsection, the analytical bounds of the eigenvalues and eigenvectors are computed.

### 2.3 The bounds of the numerical eigenvalues

As shown in Eq. (5), the interval modal parameters of the numerical model are required in order to perform the optimization procedure. Generally, the eigenvalues behave monotonically during the variation of the updating parameters if the eigenvalues are linear functions of the mass and stiffness matrices. In this case, the bounds of the eigenvalues and the eigenvectors are given by Eqs.(7) and (8):

$$\overline{\boldsymbol{K}}\,\overline{\boldsymbol{\phi}}_j = \overline{\lambda}_j \underline{\boldsymbol{M}}\overline{\boldsymbol{\phi}}_j \tag{7}$$

$$\underline{\boldsymbol{K}}\,\underline{\boldsymbol{\phi}}_j = \underline{\lambda}_j \overline{\boldsymbol{M}}\underline{\boldsymbol{\phi}}_j \tag{8}$$

More details about the computing procedure can be found in [23].

### 2.4 The ACO algorithm for continuous domain

In this paper, the metaheuristic optimization techniques are selected to perform the FFEMU. These optimization algorithms belong to the class of the population based algorithms (swarm intelligence). Generally, the metaheuristic optimization algorithms are accurate and produce reliable results. Unfortunately, some of these algorithms are computationally expensive due to the large population sizes involved in order to obtain accurate results. In this study, the ACO algorithm [24, 25] is employed since it requires less population (ants) size due to its ability to adjust to small (or large) search spaces. The results obtained by the ACO algorithm are highlighted by comparing its results with the PSO algorithm [26, 27, 28] results. The PSO algorithm was first introduced by Kennedy and Eberhart [26] where this algorithm was inspired from the behavior of flocks of birds (particles) in nature. In the standard PSO algorithm, particles (or individuals) represent potential solutions to an optimization problem where each particle is composed of three vectors: its position in the search space**,** the best position that has been found by the individual (or particle) and the velocity of the individual. Further details about the PSO algorithm can be found in [26, 27, 28].

Recently, the application of the ACO algorithm has increased in many engineering areas. This algorithm was inspired by the colonies of ants, and has been frequently used to solve both continuous and discrete optimization problems. In this section, the ACO for continuous optimization problems (sometimes called $ACO_{\mathbb{R}}$), which is proposed by Socha and Dorigo [24], is discussed in detail. In this algorithm, the probability density function (PDF) is used in the decision making process where a set of solutions (called a solution archive) is usually sampled and used to obtain the probability density function. First, the solution archive is initialized by randomly generating $Q$ solutions. Next, $P$ new solutions are generated and added to the previous $Q$ solutions so that the solution archive size becomes $Q + P$. Then, these solutions are sorted according to an objective function and the best $Q$ solutions are stored in the solution archive. This procedure is called archive update where the solution archive matrix is represented in Eq. (9).

$$\boldsymbol{T} = \begin{bmatrix} \breve{x}_1^1 & \breve{x}_1^2 & \cdots & \breve{x}_1^i & \cdots & \breve{x}_1^d \\ \breve{x}_2^1 & \breve{x}_2^d & \cdots & \breve{x}_2^d & \cdots & \breve{x}_2^d \\ \vdots & \vdots & \cdots & \vdots & \cdots & \vdots \\ \breve{x}_j^1 & \breve{x}_j^2 & \cdots & \breve{x}_j^i & \cdots & \breve{x}_j^d \\ \vdots & \vdots & \cdots & \vdots & \cdots & \vdots \\ \breve{x}_Q^1 & \breve{x}_Q^2 & \cdots & \breve{x}_Q^i & \cdots & \breve{x}_Q^d \end{bmatrix} \tag{9}$$

where $d$ represents the number of solution components or the size of the updating parameters, $Q$ is the number of solutions stored in the archive while $\breve{x}_j^i$ represents the $i$-th solution component (or the $i$-th updating parameter) of $j$-th solution. As indicated by Eq. (9), each row in the matrix $\boldsymbol{T}$ corresponds to a possible solution and the value of the objective function provides the value of each solution. The probabilistic solution construction procedure is one of the important procedures in the ACO algorithm. In this procedure, a multimodal one-dimensional probability density function based on a Gaussian kernel is employed to generate (or construct) solutions. This Gaussian kernel function consists of the weighted sum of several Gaussian functions $g_j^i$ where $i$ is the coordinate index (or the solution component) and $j$ is the solution index. The Gaussian kernel for the coordinate $i$ is given by:

$$G^i(x) = \sum_{j=1}^{Q} w_j g_j^i(x) = \sum_{j=1}^{Q} w_j \frac{1}{\sigma_j^i\sqrt{2\pi}} e^{-\frac{\left(x-\mu_j^i\right)^2}{2{\sigma_j^i}^2}} \tag{10}$$

Note that: $j \in \{1, \dots, Q\}$ and $i \in \{1, \dots, d\}$. The quantity $w_j$ represents the weight with the ranking of solution $j$ in the archive, $rank(j)$. The weight is also calculated according to a Gaussian function:

$$w_j = \frac{1}{qQ\sqrt{2\pi}} e^{-\frac{(rank(j)-1)^2}{2q^2Q^2}} \tag{11}$$

where $q$ is a constant with a small value. In the solution construction process, each ant uses a probabilistic quantity defined in Eq. (12) to select one of the solutions in the archive according to its corresponding weight.

$$p_j = \frac{w_j}{\sum_{i=1}^{Q} w_i} \tag{12}$$

Then, the new component $\check{x}_j^i$ is obtained by sampling around the selected component using a Gaussian PDF where the mean value $\mu_j^i = \check{x}_j^i$, and $\sigma_j^i$ is defined by:

$$\sigma_j^i = \xi \sum_{r=1}^{Q} \frac{\left|\check{x}_r^i - \check{x}_j^i\right|}{Q-1} \tag{13}$$

$\sigma_j^i$ represents the average distance between the $i$-th component of the solution $\check{x}_j$ and the $i$-th component of the other solutions in the archive matrix while $\xi$ is a constant. The constant $\xi$ has a similar effect to that of the pheromone evaporation rate in ACO for discrete optimization [25]. After each solution construction, the pheromone update is performed by adding the new $P$ constructed solutions to the solution archive matrix $\boldsymbol{T}$, then only the best $Q$ solutions are kept in the archive solution. The ACO algorithm will be used to minimize the objective function described in Eq. (5).

## 3. Numerical illustrations

In this subsection, a five-DOF linear system shown in Figure 1 is updated by correcting a vector of 5 unknown parameters $\boldsymbol{\theta} = \{\theta_1, \theta_2, \theta_3, \theta_4, \theta_5\}$. The system consists of 5 masses and 10 springs where the deterministic parameters of the system are: $m_1 = m_2 = 27$ kg, $m_3 = 71$ kg, $m_4 = 53$ kg, $m_5 = 29$ kg, $k_3 = 3200$ N/m, $k_5 = 1840$ N/m, $k_7 = 2200$ N/m, $k_9 = 2800$ N/m and $k_{10} = 2000$ N/m. The parameters: $k_1$, $k_2$, $k_4$, $k_6$ and $k_8$ are the uncertain parameters where their uncertainties are defined as fuzzy membership functions. First, the simulated measured data were generated using the true values of the uncertain parameters. These simulated measurements will be used to update the fuzzy membership functions of the input parameters using the proposed FFEMU based ACO algorithm. The results obtained by the FFEMU based ACO algorithm are compared with the results obtained by the FFEMU based PSO algorithm and the Metropolis-Hastings (M-H) algorithm.

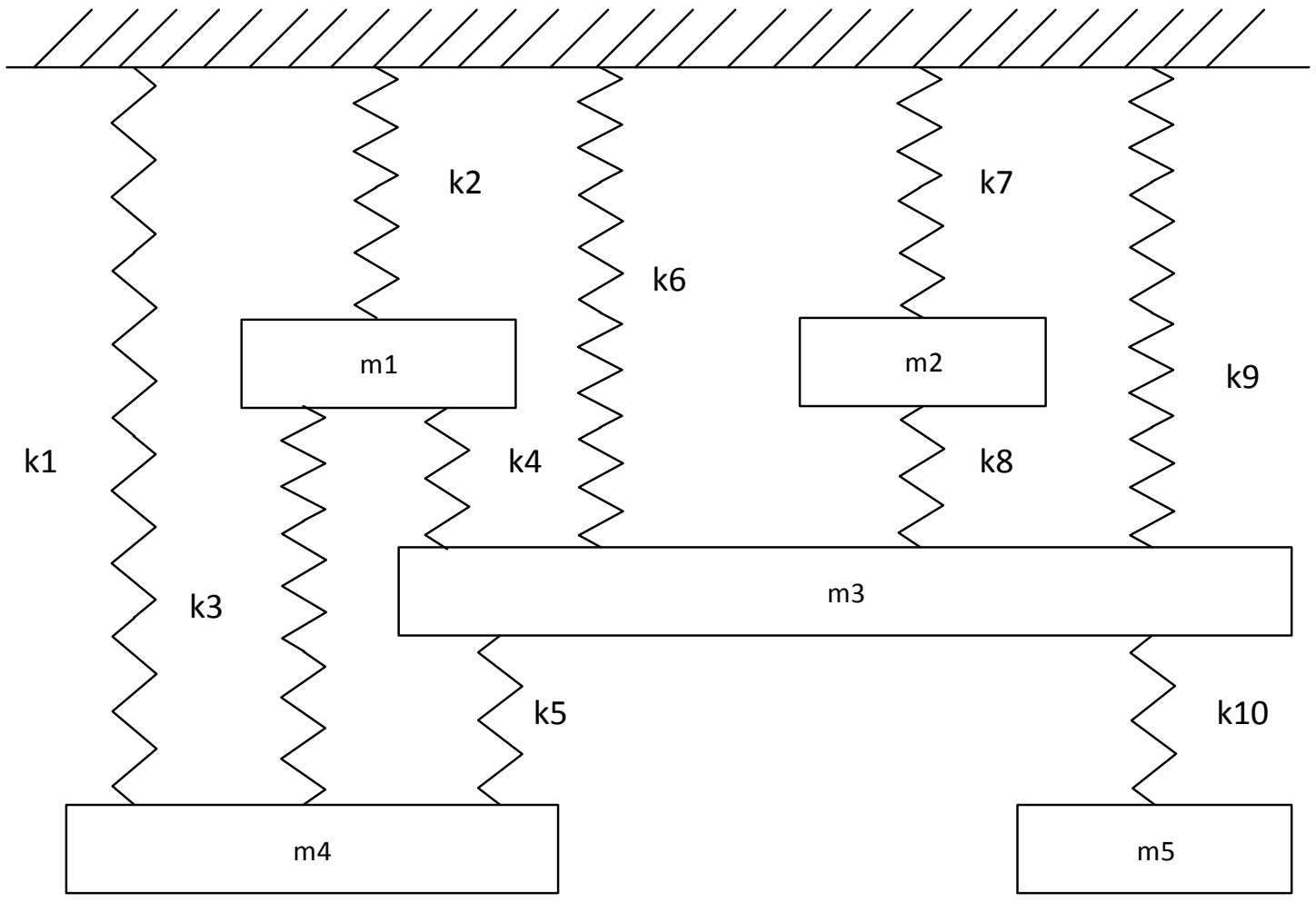

Figure 1: Five degree of freedom mass-spring system

In the FFEMU algorithms, 10 α-cuts were used in the optimization procedure. The general bounds $\boldsymbol{\theta}_{max}$ and $\boldsymbol{\theta}_{min}$ are set to: $\{4600,2500,2370,3300,2650\}$ and $\{3400,1900,1700,1900,2150\}$, respectively. For the ACO algorithm, the value of $Q$ is set to 10, $P = 20$, $q = 0.5$ and $\xi = 1$. Concerning the PSO algorithms, 80 particles were used for the optimization while the number of samples is set to $N_s = 10000$ for the M-H algorithm. The membership functions of the uncertain parameters obtained by the FFEMU based ACO and the FFEMU based PSO algorithms are shown in Figures 2 and 3, respectively. The density functions obtained by the M-H algorithm are shown in Figure 4.

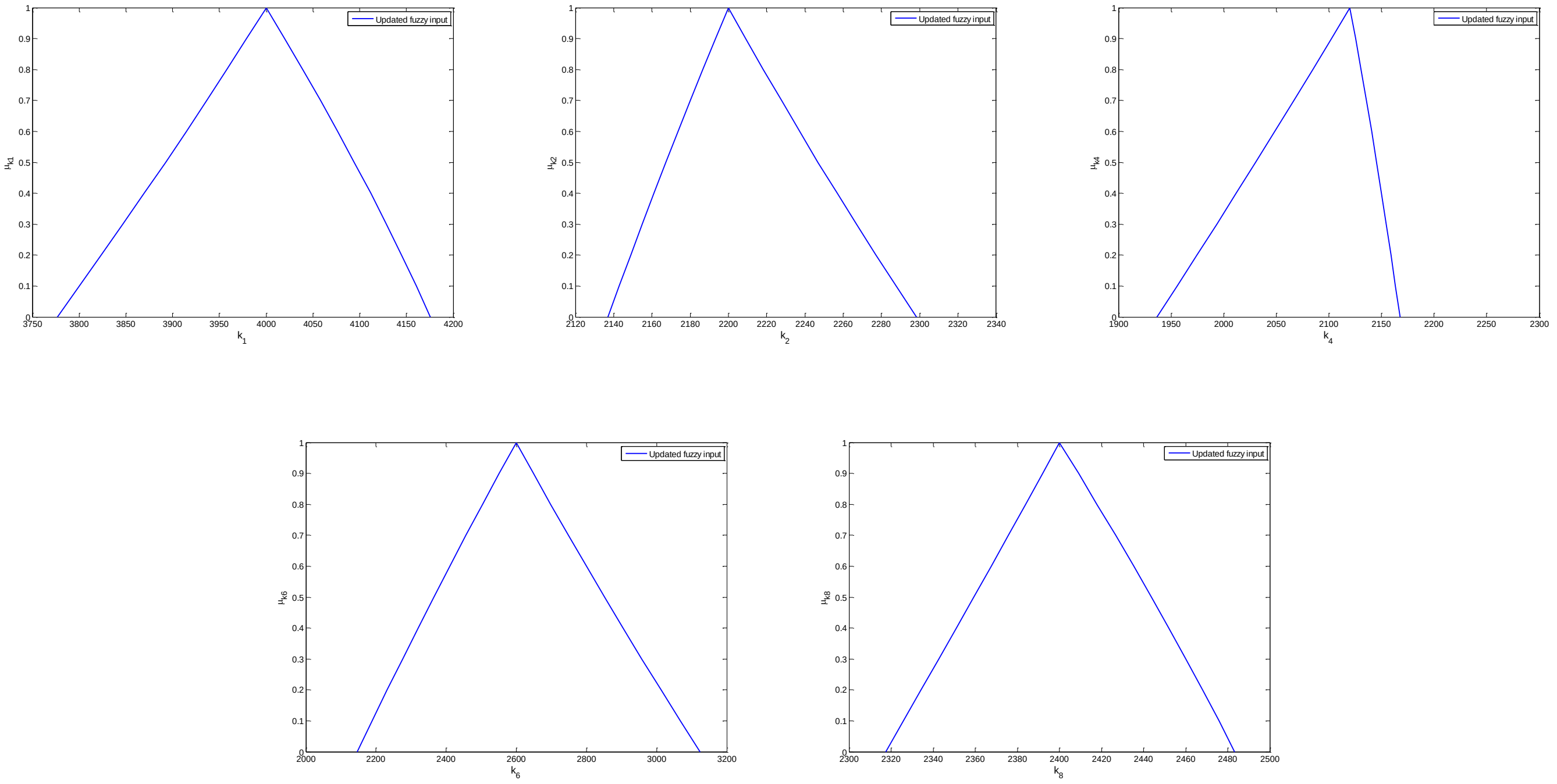

Figure 2: The membership functions of the updating parameters obtained by the FFEMU base ACO algorithm

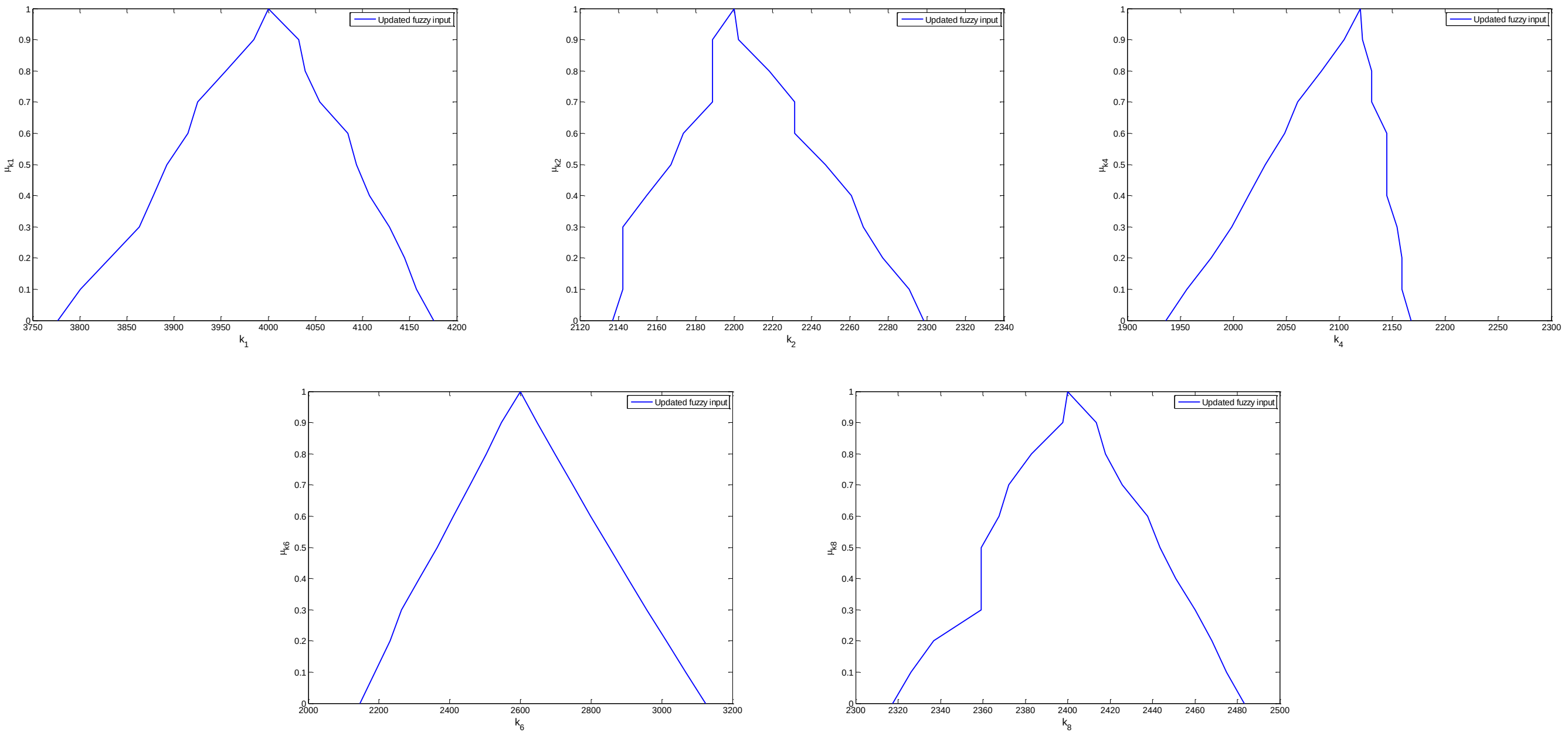

Figure 3: The membership functions of the updating parameters obtained by the FFEMU base PSO algorithm

As shown in Figures 2 and 3, the membership functions obtained by the FFEMU based ACO algorithm have perfect triangular forms which are not the case with the membership functions obtained by the FFEMU based PSO algorithm. Moreover, the FFEMU based ACO algorithm does not require a large number of ants $Q$ while the number of particles used by FFEMU based PSO algorithm is 80 particles (the FFEMU based PSO algorithm gives poor membership forms when less than 80 particles are involved in the optimization). The reason that the FFEMU based ACO algorithm give better shapes with fewer ants is that the solution construction procedure is fully automated (see Eq. (13)) while the PSO algorithm uses the velocity bounds to generate the new velocity of the particles (to generate new solutions). These bounds of the velocity are fixed and when the algorithm moves from one interval to the next, the search space will be reduced according to the constraints presented in Eq. (6). The new search space may require a different velocity bounds to obtain better results which is not true since the velocity bounds are already fixed. To overcome this weakness, a large number of particles are used to ensure good results or the velocity bounds are adjusted in every α-cut. The density functions presented in Figure 4 show some similarity with the membership functions in term of variation range of the parameters.

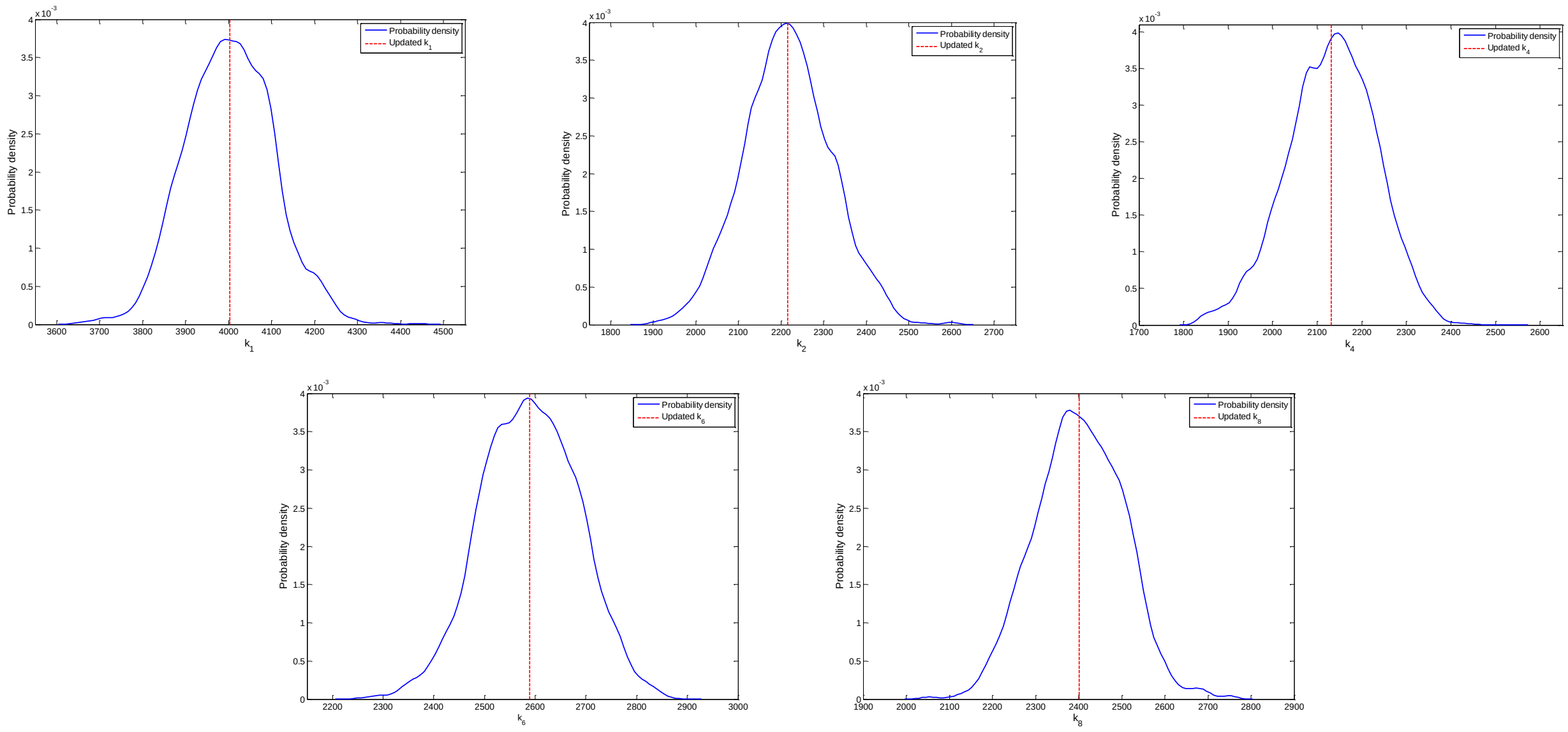

Figure 4: The probability density functions of the updating parameters obtained by the M-H algorithm

The updating values of the uncertain parameters are presented in Table 1 along with their initial values, measured values and the updating bounds. In the case of M-H algorithm, the coefficients of variation (c.o.v) values are also presented. Table 1 shows that the updating parameters along with their bounds obtained by FFEMU based ACO and FFEMU based PSO algorithms are exactly the same. The results obtained by the M-H algorithm are slightly different which was expected since the PSO and ACO algorithms are more accurate than the Monte Carlo methods. The c.o.v values obtained by the M-H algorithm are relatively small (less than 5%) which indicates that the M-H algorithm has efficiently estimated the uncertain parameters.

Table 1: Initial, intervals and updated parameters when FFEMU and M-H algorithms are used to update the structure

| | $\boldsymbol{\theta_0}$ vector Initial | $\boldsymbol{\theta}$ vector, FFEMU-ACO Method | $\boldsymbol{\theta}$ interval, FFEMU-ACO Method | $\boldsymbol{\theta}$ vector, FFEMU-PSO Method | $\boldsymbol{\theta}$ interval, FFEMU-PSO Method | $\boldsymbol{\theta}$ vector, M-H Method | $\frac{\sigma_i}{\theta_i}$ (%) |
|---|---|---|---|---|---|---|---|
| $k1$ | 4150 | 4000 | [3776.4, 4175.6] | 4000 | [3776.4, 4175.6] | 4002.6 | 2.55% |
| $k2$ | 2150 | 2200 | [2137.0, 2298.3] | 2200 | [2137.0, 2298.3] | 2215.5 | 4.61% |
| $k4$ | 2160 | 2120 | [1936.6, 2167.6] | 2120 | [1936.6, 2167.6] | 2132.0 | 4.63% |
| $k6$ | 2500 | 2600 | [2146.7, 3123.9] | 2600 | [2146.7, 3123.9] | 2589.3 | 3.66% |
| $k8$ | 2460 | 2400 | [2317.4, 2483.4] | 2400 | [2317.4, 2483.4] | 2401.7 | 4.24% |

Figures 5 and 6 show the membership functions of the updated outputs for the FFEMU based ACO and FFEMU based PSO algorithms, respectively. The FFEMU based ACO algorithm gives an excellent match between fuzzy membership functions of the updated outputs and the measured outputs. Also, the membership functions obtained by the FFEMU based ACO are slightly better than those obtained by the FFEMU based PSO algorithm.

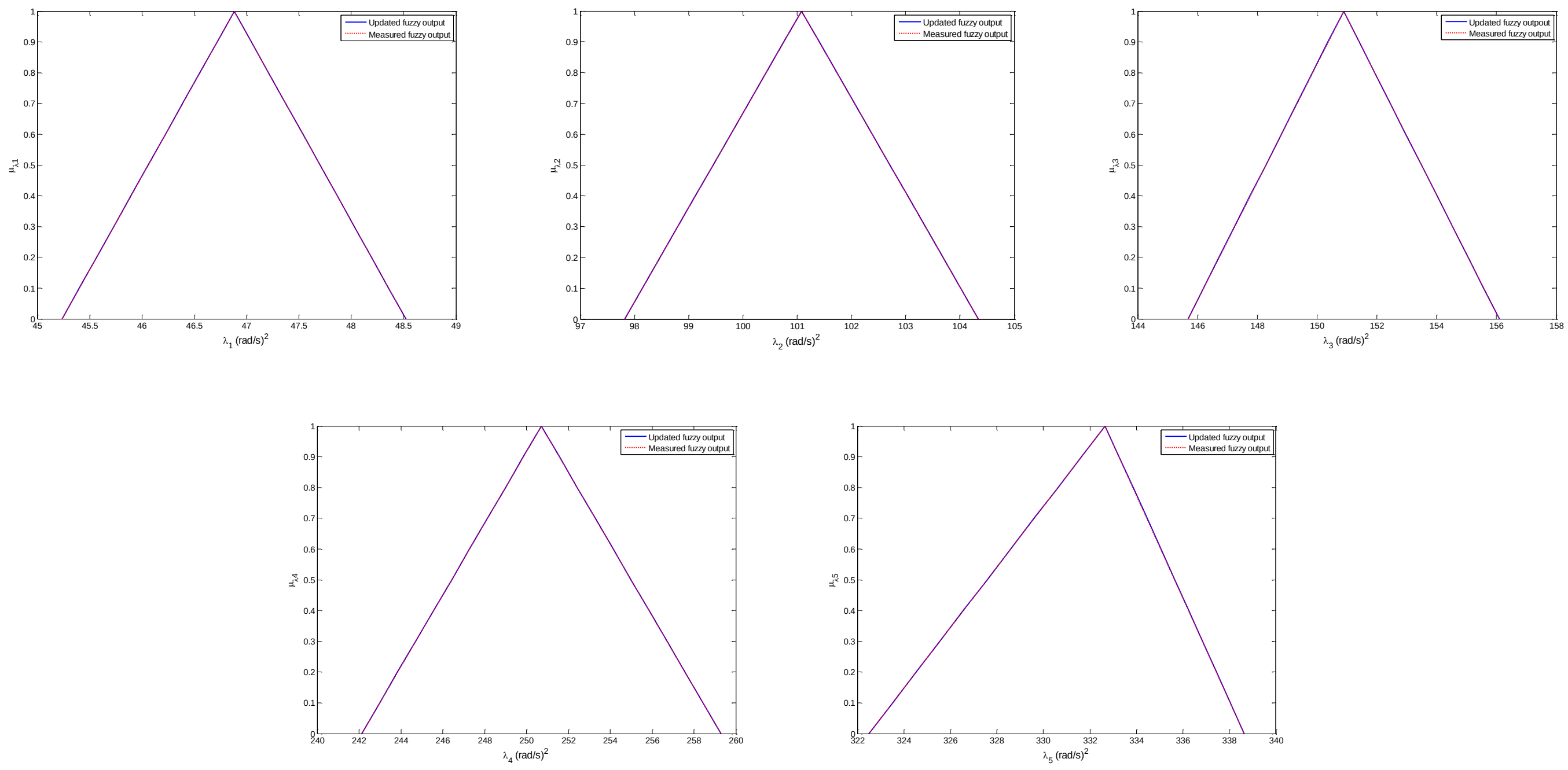

Figure 5: The membership functions of the updating outputs obtained by the FFEMU base ACO algorithm

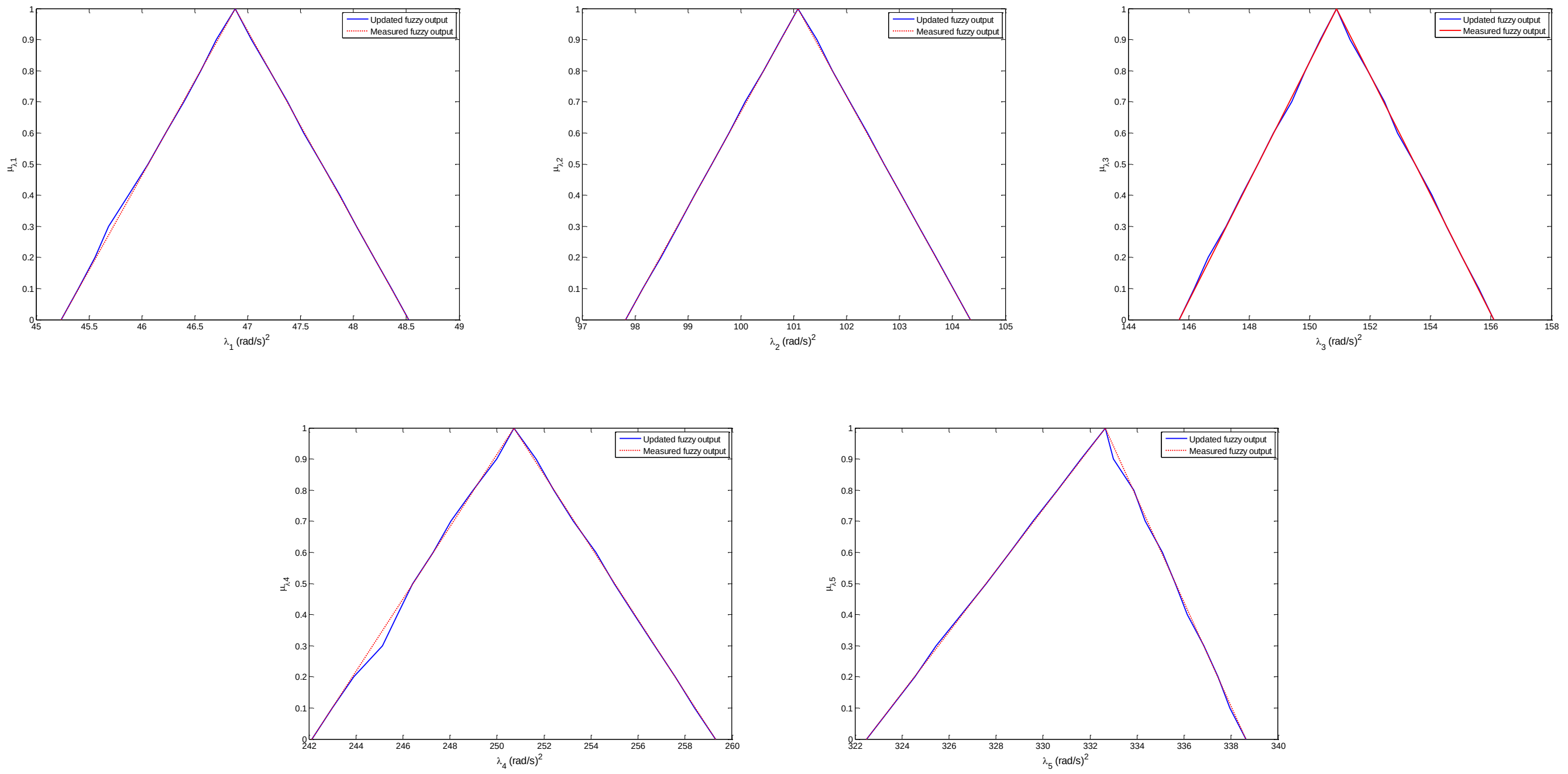

Figure 6: The membership functions of the updating outputs obtained by the FFEMU base PSO algorithm

Table 2 shows the initial, updated eigenvalues and interval eigenvalues along with the absolute errors obtained by all methods. Again, the updating eigenvalues and their bounds obtained by both FFEMU based ACO and FFEMU based PSO algorithms are exactly the same. The M-H algorithm gives also good results in term of total error (0.17%) or c.o.v values (less than 2 %).

Table 2: The eigenvalues, intervals and errors when FFEMU and M-H algorithms are used to update the structure

| Mode | Measured eigenvalues (Hz) | Initial eigenvalues (Hz) | Error (%) | Eigenvalues FFEMU-ACO Method (Hz) | Error (%) | Interval FFEMU-ACO Method (Hz) | Eigenvalues FFEMU-PSO Method (Hz) | Error (%) | Interval FFEMU-ACO Method (Hz) | Eigenvalues M-H Method (Hz) | Error (%) |
|---|---|---|---|---|---|---|---|---|---|---|---|
| 1 | 47.81 | 46.50 | 1.99 | 46.88 | 0.00 | [45.24, 48.52] | 46.88 | 0.00 | [45.24, 48.52] | 46.90 (0.73%) | 0.03 |
| 2 | 104.05 | 100.87 | 2.94 | 101.08 | 0.00 | [97.82, 104.34] | 101.08 | 0.00 | [97.82, 104.34] | 101.23 (1.17%) | 0.15 |
| 3 | 155.31 | 148.52 | 2.93 | 150.89 | 0.00 | [145.68, 156.10] | 150.89 | 0.00 | [145.68, 156.10] | 151.36 (1.91%) | 0.31 |
| 4 | 254.52 | 248.08 | 1.52 | 250.72 | 0.00 | [242.13, 259.29] | 250.72 | 0.00 | [242.13, 259.29] | 250.91 (1.48%) | 0.08 |
| 5 | 342.71 | 333.89 | 3.02 | 332.66 | 0.00 | [322.50, 338.64] | 332.66 | 0.00 | [322.50, 338.64] | 333.57 (1.38%) | 0.28 |
| Total average error | ________ | ________ | **2.48** | _______ | **0.00** | _______ | _______ | **0.00** | _______ | _______ | **0.17** |

## 4. Conclusion

In this paper, an FFEMU algorithm was implemented to identify the membership functions of the inputs. The FFEMU problem was formulated as an optimization problem where the ACO algorithm was employed to provide the membership functions of the inputs. To test this algorithm, a five-DOF mass-spring system was updated using the FFEMU based ACO algorithm. The results obtained were also compared with FFEMU based PSO and M-H algorithms results. The ACO algorithm was able to obtain very good results with few ants. In further work, more complex structures will be investigated and the fuzzy finite element model will be improved in term of computational cost and accuracy.